\documentclass[sigconf]{acmart}

\AtBeginDocument{%
  }

\setcopyright{acmcopyright}
\copyrightyear{2022}
\acmYear{2022}
\acmDOI{XXXXXXX.XXXXXXX}

\acmConference[Conference acronym 'XX]{Make sure to enter the correct
  conference title from your rights confirmation emai}{June 03--05,
  2018}{Woodstock, NY}
\acmPrice{15.00}
\acmISBN{978-1-4503-XXXX-X/18/06}

\usepackage{adjustbox}
\usepackage{graphicx}
\usepackage{subfig}
\usepackage{enumitem}

\begin{document}

\title{Document Summarization with Text Segmentation}

\author{Lesly Miculicich}
\email{leslym@microsoft.com}
\affiliation{%
  \institution{Microsoft AI}
  \city{Bellevue}
  \state{Washinton}
  \country{USA}
}
\orcid{1234-5678-9012}
\author{Benjamin Han}
\authornotemark[1]
\email{dingjung.han@microsoft.com}
\affiliation{%
  \institution{Microsoft AI}
  \city{Bellevue}
  \state{Washinton}
  \country{USA}
}


\begin{abstract}
In this paper, we exploit the innate document segment structure for improving the extractive summarization task. We build two text segmentation models and find the most optimal strategy to introduce their output predictions in an extractive summarization model. Experimental results on a corpus of scientific articles show that extractive summarization benefits from using a highly accurate segmentation method. In particular, most of the improvement is in documents where the most relevant information is not at the beginning thus, we conclude that segmentation helps in reducing the lead bias problem.

\end{abstract}



\maketitle

\section{Introduction}
Extractive summarization offers the ability to retrieve relevant sentences in a document. The applications of this technology can speed up organization's work and could serve as pre-processing step for other document understanding tasks in which the input is too long to be easily processed. 

Documents have an internal structure that can be explicit like content tables, or implicit like change of topics. The exploitation of these structures is pertinent for summarization for helping to locate the most relevant information in the document. Thus our objective is to automatically detect text segments and incorporate this information in an extractive summarization model to boost its accuracy. Given that the automatic extraction of document structures have challenges in real application scenarios, we aim at detecting implicit changes of topics in the text. 

We explore two state-of-the art models for text segmentation, one trained with supervised learning and the other is unsupervised. Supervised methods exhibit better accuracy however are limited by the feature characteristics of the training data such as domain, genre, and distribution; thus they have difficulties for generalizing to unseen data. Unsupervised methods work in a more generic manner at the cost of less correctness over in-domain data. We use the prediction of these methods in an extractive summarization model. We assess a comprehensive set of strategies for integrating segment data in the summarization model, and show that segmentation increases the quality of the summarization model on a corpus of scientific papers \cite{cohan-etal-2018-discourse}. However, the effectiveness depends on the accuracy of the segmentation model. 

This paper is organized as follows: In Section~\ref{sec:bck}, we described the related work; in Sections~\ref{sec:seg}~and~\ref{sec:sum}, we define our models for text segmentation and summarization respectively; in Section\ref{sec:data}, we report the dataset and metrics used for the evaluation and the results are shown in Section~\ref{sec:exp}. Finally, Section~\ref{sec:conclu} contains our conclusions and future work.

\section{Related work}\label{sec:bck}

On unsupervised segmenation, one of the first successful approaches was TextTiling \cite{hearst-1997-text}. It utilizes similarity scores between adjacent sentences to decide whether there is a change of segments. The original work uses sentence similarity based on word frequency. Later work updated this method by using cosine similarity between sentence embeddings \cite{xu2021topicaware, solbiati2021unsupervised}. On supervised segmentation, the later work uses sequence-to-sequence learning, for instance \cite{liu-etal-2022-end} uses a sequence encoder and, for each token, predicts whether there is a new segment or not. In \cite{ijcai2018-579}, the authors use an RNN based pointer-network, the RNN has as many time-steps as segments, at each time step, it selects a token which indicates the end of a segment. 

Text segmentation has being successfully applied to automatic summarization. Most of this work is evaluated on meeting transcripts. The main idea is to split the documents on segments and then summarize each segment. This can be implemented in a pipeline (first segment then summarize) or in a end-to-end fashion \cite{liu-etal-2022-end, liu2019topic}. More related to the present work, \cite{ruan-etal-2022-histruct} proposed an extractive summarization model that includes information for segments. However, they use \emph{Oracle} document sections, here, we train a segmentation model to predict the sections and then we integrate this information in the summarization model. 

\section{Text Segmentation}\label{sec:seg}
Text segmentation is the task of dividing a text in meaningful parts such topics or sections. In this work, we present an unsupervised and a supervised method for document segmentation. Both approaches have state-of-the art results on topic segmentation for meeting transcripts. Here, we adapt and optimize them for section detection in text documents. 
The task is defined as follows: Given a set of input sentences $S = \{S_1, …, S_M\}$ with an underlying segment structure; the objective is to predict a sequence $Y = \{y_1, …, y_M\}$ where $y_i$ is a binary value indicating whether $S_i$ is the beginning of a new segment. 

\subsection{Unsupervised Segmentation}
We based our model on the unsupervised segmentation approach proposed in \cite{solbiati2021unsupervised}. It is a modified version of TextTiling \cite{hearst-1997-text} that detects topic changes with a similarity score based on BERT embeddings \cite{devlin-etal-2019-bert}. We first compute the representations for every sentence. Then, we divide the document in overlapping windows and perform max pooling to get the window representation. We compute cosine similarity among adjacent windows and derive segment boundaries where the semantic similarity is lower than a given threshold.
We adjust the window size, and the similarity threshold parameters using a validation set.

\subsection{Supervised Segmentation}
We use the supervised segmentation approach proposed in \cite{zhang2021sequence}. We divide the document in overlapping windows, and each window is encoded with a transformer network \cite{10.5555/3295222.3295349}. The segmentation is performed as a sequence labeling task, where each token in the sequence is assigned with a binary label to indicate weather it is the start of a new segment. We initialize the weights of the model with the pretrained model DeltaLM \cite{ma2021deltalm}, and adjust the window size, stride size, and the classification threshold parameters using a validation set.

\section{Extractive Summarization}\label{sec:sum}
Extractive summarization is the task of finding the subset of sentences in a document that best summarize it. Following \cite{liu-lapata-2019-text}, we define extractive summarization as a sequence labeling task.  Given a set of input sentences $S = \{S_1, …, S_M\}$ the objective is to assign a label $y_i \in \{0, 1\}$ to each $S_i$, indicating whether the sentence should be included in the summary. The model is composed by two transformer encoders: word-level and sentence-level. Each document is tokenized and encoded with the word-level transformer. We introduce a special to token $[CLS]$ at the end of each sentence. The output vector corresponding to  this token serves as the sentence representation. All sentence representations plus their corresponding positions are input to a secondary sentence-level transformer. The word-level transformer is initialized with the pretrained model DeltaLM \cite{ma2021deltalm} whereas the sentence-level transformer is initialized with random values, and it is composed of only 2 layers. We use binary-cross entropy to train the model. 

In order to manage long input sequences, the documents are chunked into equal size block. Each chunk is encoded with the word-level transformer independently. Then the output of the chunks are concatenated and the rest of the model is the same. 

\subsection{Integrating Segment Information}
We deem adequate to integrate the segment information in the sentence-level transformer. The referred segment information can be either the segment position in the document or the segment semantic representation. Both are relevant and serve different purposes, one learns the location of the relevant information, and the other spots the relevant segment depending on its content. 

\subsubsection{Segment position encoding} 
We use a learned positional encoding with a maximum of 10 segments. To avoid position bias, we applied normalization to the number of segments in the document as follows: $pos_i = I*max_{seg}/(n_{seg} + 1)$, where $pos_i$ is the final position, $i$ is the segment index, $max_{seg}=10$, and $n_{seg}$ is the number of segments in the document. 

\subsubsection{Segment Embedding}
We calculate the segment embedding by applying pooling to its tokens embeddings. We used maximum, minimum, and mean pooling. Preliminary experiments showed that mean pooling has better results, thus we use it.

\subsubsection{Segment Position HiStruct} 
Following \cite{ruan-etal-2022-histruct}, we encode positions in a hierarchical manner by summing the segment position and sentence relative position in the corresponding segment. According the original experiments in the paper, using learnable embeddings and summing performed the best, thus we apply the same strategy.

We integrated the segment information by either adding or concatenation the segment representations.


\section{Data and Metrics}\label{sec:data}
For our experiments, we use Arxiv dataset \cite{cohan-etal-2018-discourse}. It contains scientific articles with annotation of sections and sentences. It is composed of 203,037 samples for training, 6,436 for validation, and 6,440 for testing. We use sections as segment markers. Table~\ref{tab:arxiv} shows the statistics of sentences per section in the training set.

\begin{table}
	\centering
	\begin{tabular}{l  c } 
	\toprule
	\textbf{Statistics} & \textbf{$\#$}\\ \hline
    Avg. sentences per summary&	11\\
    Avg. sections per document&	5.5\\
    Avg. sentences per document&	130\\
    Max. sentences per summary&	110\\
    Max. sections per document&	60\\
    Max. sentences per document&	1268\\
    \bottomrule
   \end{tabular} 
	\caption{Sections and sentences statistics on Arxiv data-set.}
	\label{tab:arxiv}
\end{table}

\subsection{Summarization} We use ROUGE score \cite{lin-2004-rouge} for evaluating the summarization models. It measures the n-gram overlapping between the predicted summary and a reference. We report the F1 score of uni-grams (R1), bi-grams (R2), and the longest matching sequence (RL).

\subsection{Text Segmentation} Two standard evaluation metrics are used to evaluate text segmentation: Pk \cite{beeferman1999statistical} and WinDiff \cite{pevzner-hearst-2002-critique}. Pk represents the probability that a randomly chosen pair of words at a distance of $k$ is inconsistently classified; that is, for one segmentation the pair lies within the same segment, while for another the pair spans across segment boundaries. This is implemented by using a sliding window of size set to half of the average true segment length, and counting how many times the predictions differ from the reference. This probability can be further decomposed into two conditional probabilities: the miss and the false alarm probabilities. WinDiff is a modification of Pk where the algorithm slides a fixed-sized window across the text and penalizes whenever the number of predicted boundaries within the window does not match the true number of boundaries within the same window.

\subsection{Lead bias} \label{sec:lead} Following \cite{grenander-etal-2019-countering}, we calculate the R1 score on three different label distributions – D-early, D-middle and D-last – which are obtained by first sorting documents by the average sentence position of the positive labeled sentences: D-early are the first 100 documents, D-middle are the middle 100 documents, and D-late are the last 100 documents.

\section{Experimental Results}\label{sec:exp}
In this section, we describe the results of both unsupervised and supervised text segmentation methods. We also compare the different methods for integrating text segmentation on the extractive summarization model based on ground truth segments. Finally, we use the best integration method to report final results on summarization.

\subsection{Unsupervised Segmentation}
We tuned the hyper-parameters of the model using 15 documents from the validation set. We evaluate a window size in the range $[1,5,10,15]$ and threshold in $[0.4,0.5,0.6]$ (see Table~\ref{tab:results:unsup}). We picked 0.5 and 5 as the threshold and the window size respectively. 

\begin{table}
	\centering
	\begin{adjustbox}{width=\linewidth} 
	\begin{tabular}{l  c c | c c | c c} 
		  & \multicolumn{6}{c}{\textbf{Threshold}} \\ \hline
		  & \multicolumn{2}{c |}{0.4} &	\multicolumn{2}{c|}{\textbf{0.5}} & \multicolumn{2}{c}{0.6} \\ \hline
        \textbf{Window size} &	Pk &	WinDiff&	Pk&	WinDiff	&Pk&	WinDiff\\ \hline
        1&	0.491&	0.535&	0.469&	0.506&	0.494&	0.509\\
        \textbf{5}&	0.419&	0.483&	\textbf{0.418}&	\textbf{0.461}&	0.455&	0.482\\
        10&	0.469&	0.531&	0.476&	0.514&	0.492&	0.537\\
        15&	-- & --	 &	-- &--	 &	0.490&	0.516 \\ \hline
    \end{tabular} 
	\end{adjustbox}
	\caption{Hyper-parameter tuning for the unsupervised text segmentation model.}
	\label{tab:results:unsup}
\end{table}

We include two simple baselines models: a \emph{Random} method that places segment boundaries uniformly at random, and an \emph{Even} method that places boundaries every $k$ sentences. The results are shown in Table~\ref{tab:results:unsup2}.

\begin{table}
	\centering
	\begin{tabular}{l c c } 
		\toprule
		\textbf{Model} & \textbf{Pk}	&  \textbf{WinDiff} \\ \hline
        Random&	0.544&	0.703\\
        Even&	0.503&	0.516\\
        Unsupervised segmentation & 0.403&	0.437\\ 
        Supervised segmentation & \textbf{0.183} &	\textbf{0.224}\\
        \bottomrule
    \end{tabular} 
	\caption{Comparison of text segmentation methods.}
	\label{tab:results:unsup2}
\end{table}

\begin{figure*}[ht]
   \subfloat[]{%
      \includegraphics[clip, width=0.3\textwidth]{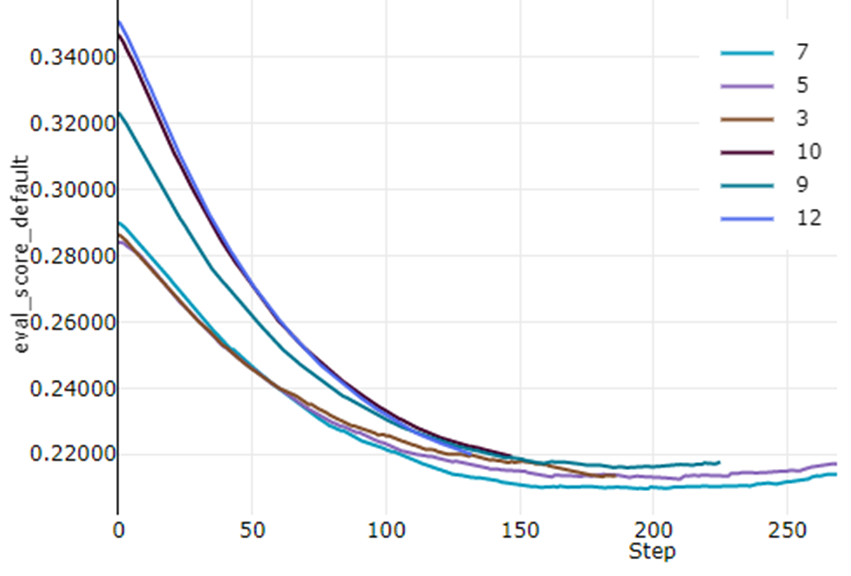}}
 \hspace{\fill}
   \subfloat[ ]{%
      \includegraphics[clip, width=0.3\textwidth]{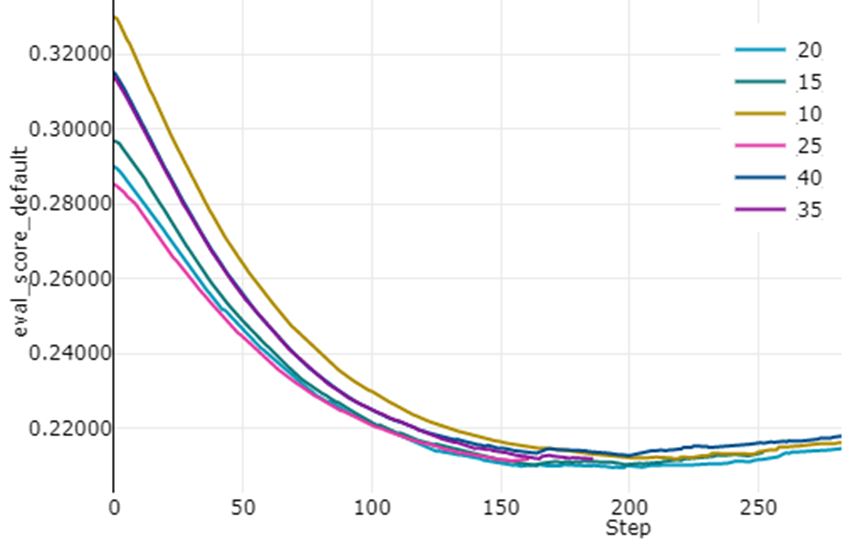}}
    \hspace{\fill}
   \subfloat[]{%
      \includegraphics[clip, width=0.3\textwidth]{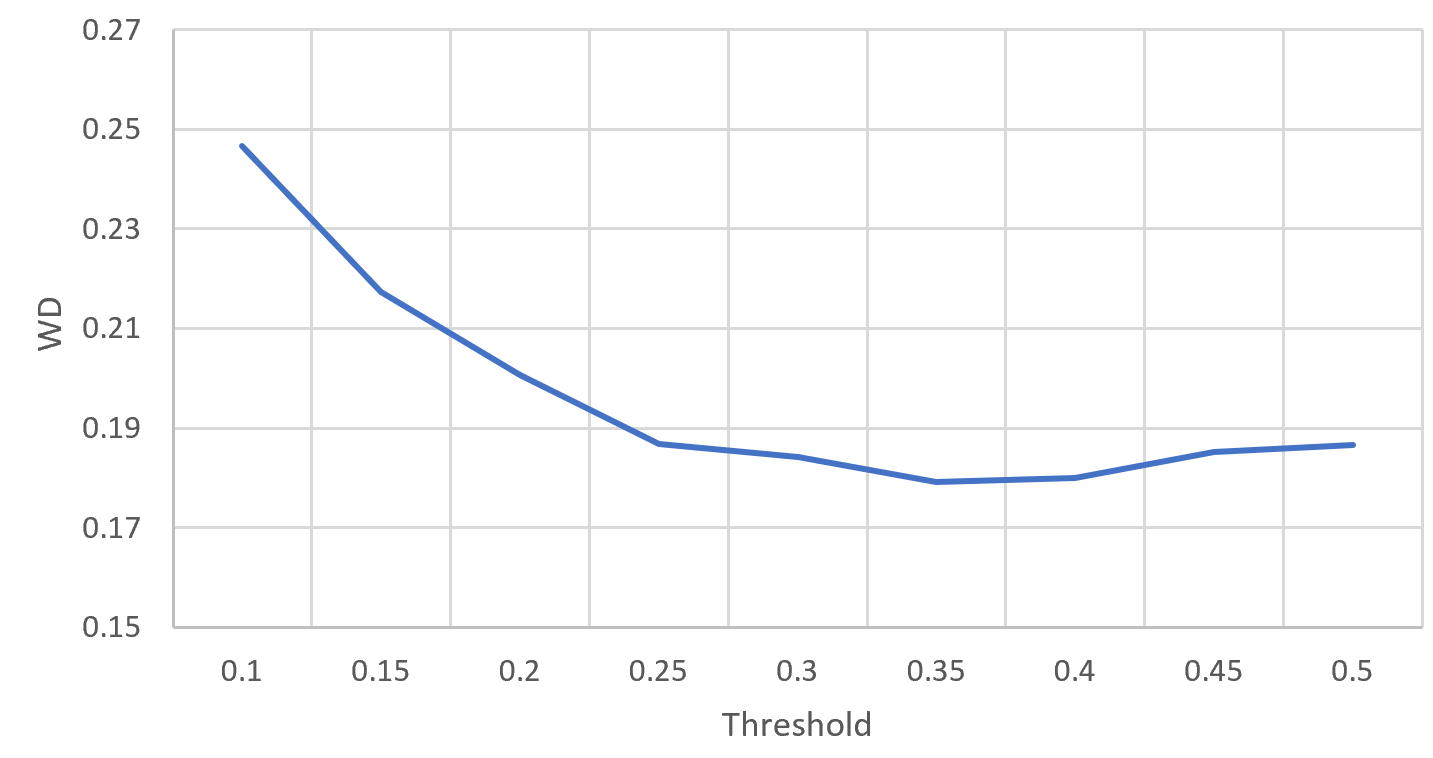}}\\
\caption{Hyper-parameter tuning for the supervised segmentation model. (a) Window; (b) Stride; (c) Threshold.}
\label{sup}
\end{figure*}

\subsection{Supervised Segmentation}
In the supervised segmentation model, a document is processed by sliding windows. The widow’s size is the number of sentences to be processed in a chunk. We tested sizes in the range of $[10, 40]$. Figure~\ref{sup}(a) shows the Pk score in the development set. The best score was obtained with a window of 20. Similarly, the stride is the number of overlapping sentences in the windows. We evaluated values in the range of $[3, 12]$. Figure~\ref{sup}(b) shows the Pk score in the development set with a window size of 20. The best score was obtained with a stride of 7, the best value is consistent for all window sizes. Finally, the model outputs a score in the range of $[0,1]$ for each sentence. The score threshold determines whether the sentence is the start of a new segment. Figure~\ref{sup}(c) shows the WinDiff scores for different thresholds on the development set obtained with the best model. The optimum threshold value is 0.35. The final results are shown in Table~\ref{tab:results:unsup2}.

\subsection{Extractive Summarization with Segmentation}
Table~\ref{tab:results:int} shows the ROUGE scores of different methods of integrating segmentation information on the extractivw summarization model (ExtSum). As the objective is to evaluate the best integration method, we use the ground truth annotation for segments, named \emph{Oracle}. We found that both segment position and segment embedding help to improve the model. We also found that concatenating the segment information to the sentences input works better than adding it. Finally, using a flat position embedding have equal or slightly better results that using a hierarchical position as purposed in \cite{ruan-etal-2022-histruct}.

\begin{table}
	\centering
	\begin{adjustbox}{width=\linewidth}
	\begin{tabular}{l c c c c} 
		\toprule
		\textbf{Model} & \textbf{Op.} & \textbf{R1} &  \textbf{R2} &  \textbf{RL} \\ \hline
            ExtSum	 & &	48.91	&20.62	&43.85\\
            \quad Seg. Position &	Add.	&49.28	&20.86	&44.18\\
            \quad Seg. Embedding &	Add.	&49.01	&20.68	&43.93\\
            \quad Seg. Pos. + Embed. & 	Add.	&49.25	&20.89	&44.14\\
            \quad Seg. Pos. + Embed. 	& Concat.	&\textbf{49.49}	& \textbf{21.04}	& \textbf{44.34}\\
            \quad Seg. Pos. HiStruct + Embed.		& Concat.	&49.46	&21.01	&44.31\\ \hline
    \end{tabular} 
	\end{adjustbox}
	\caption{Comparison of methods to integrate segmentation information in the extractive summarization model.}
	\label{tab:results:int}
\end{table}

\begin{table*}
	\centering
	\begin{adjustbox}{width=\linewidth}
	\begin{tabular}{l c c c | c c c} 
		\toprule
		\textbf{Model} & \textbf{R1} &  \textbf{R2} &  \textbf{RL} & \textbf{D-early R1}  & \textbf{D-middle R1} & \textbf{D-late R1} \\ \hline	
		HiStruct$+$ (with Longformer-base 28k tok.) \cite{ruan-etal-2022-histruct} & 45.22 & 17.67 & 40.16 & -- & -- & -- \\ \hline
        ExtSum (with DeltaLM and chunked input 25K tok.) &	48.91	&20.62 &43.85 & 48.9&	55.9&	47.6\\
        ExtSum + unsupervised segmentation	& 48.63&	20.41&	43.55 & 48.5 & 55.8 & 47.4 \\
        ExtSum + supervised segmentation	&49.11 (+0.20)	&20.68 (+0.06)	&44.01 (+0.16)    &49.1 (+0.2)	&56.5 (+0.6)	&48.0 (+0.4) \\
        ExtSum + oracle segmentation	&49.49 (+0.68)	&21.04 (+0.42)	&44.34 (+0.49)   &49.4 (+0.5)	&56.8 (+0.9)	&48.1 (+0.5) \\
		\bottomrule
	\end{tabular} 
	\end{adjustbox}
	\caption{Evaluation results of using text segmentation in extractive summarization on Arxiv test-set}
	\label{tab:results:extsum}
\end{table*}

\begin{figure}
  \centering
  \includegraphics[width=\linewidth]{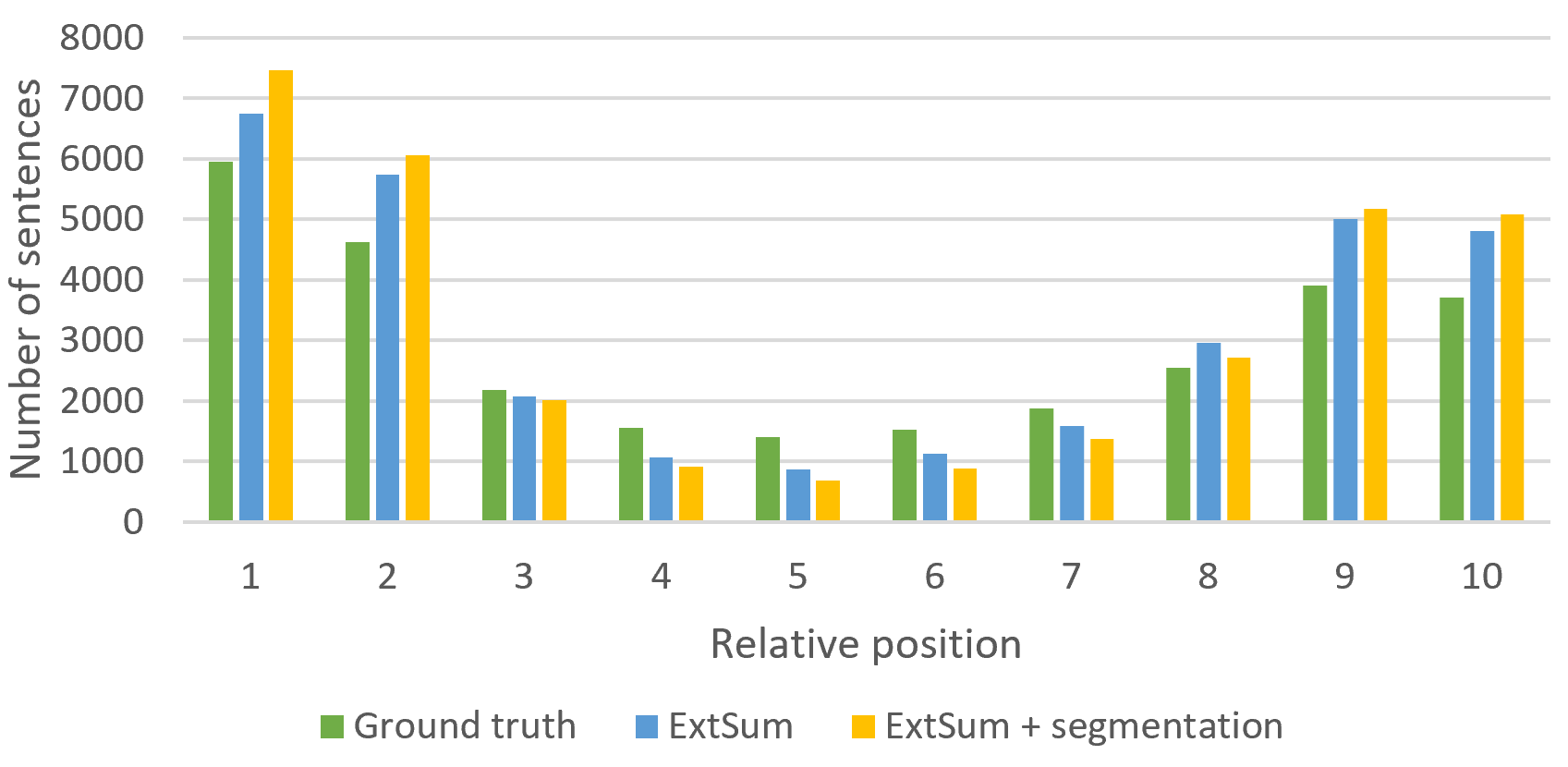}
  \caption{ Relative position of summary sentences in the document.}
  \label{fig:relpos} 
\end{figure}

We compare the results of unsupervised and supervised methods with current SOTA  \cite{ruan-etal-2022-histruct} (Table~\ref{tab:results:extsum}). We use the best model: ExtSum with contenation of segment position and embedding. We also include the results of ExtSum with \emph{oracle} segmentation to define the upper limit of this method. We measure the ROUGE scores  together with the the lead bias metrics \emph{D-early}, \emph{D-middle} and \emph{D-late} described in Section~\ref{sec:lead}. The unsupervised method rather decrease the scores for summarization. The low accuracy of the unsupervised segmentation could be adding noise instead of helping.  The supervised segmentation method show improvement above all metrics, in particular in the middle and late part of the document as shown by \emph{D-middle} and \emph{D-late}. To further analyze this results, we plot the relative position of extracted sentences in the document (Figure~\ref{fig:relpos}). We can see that adding segmentation information makes the ExtSum model extract less sentences from the middle part, and more from the early and late parts. This is consistent with Arxiv dataset, where the introduction and conclusions contains the most relevant information for the summary.

\section{Conclusions}\label{sec:conclu}
In this papers we evaluated two text segmentation methods to detect sections in documents. One unsupervised method that uses similarity scores between adjacent text blocks, and a supervised method that detects sentence by sentence when a new section starts. We combine this predictions in a extractive summarization model to boost its accuracy. We evaluated a series of strategies to combine the segment information on summarization . We show that the supervised segment predictions improve the ROUGE scores of the summarization. According our analysis, segmentation helps to detect  the sentences of most relevant sections in the dataset: introduction and conclusions (at beginning and end). However, the maximum improvement we can get using this method is about 1 point ROUGE score, as shown by our experiments using \emph{Oracle} sections. This improvement is significant but limited. Future work could include more elaborate information from the document, like hierarchical structure, section titles, and discourse information. Also, a end-to-end method to combine both summarization and segmentation.


\bibliographystyle{ACM-Reference-Format}
\bibliography{refer}


\begin{thebibliography}{17}


\ifx \showCODEN    \undefined \def \showCODEN     #1{\unskip}     \fi
\ifx \showDOI      \undefined \def \showDOI       #1{#1}\fi
\ifx \showISBNx    \undefined \def \showISBNx     #1{\unskip}     \fi
\ifx \showISBNxiii \undefined \def \showISBNxiii  #1{\unskip}     \fi
\ifx \showISSN     \undefined \def \showISSN      #1{\unskip}     \fi
\ifx \showLCCN     \undefined \def \showLCCN      #1{\unskip}     \fi
\ifx \shownote     \undefined \def \shownote      #1{#1}          \fi
\ifx \showarticletitle \undefined \def \showarticletitle #1{#1}   \fi
\ifx \showURL      \undefined \def \showURL       {\relax}        \fi
\providecommand\bibfield[2]{#2}
\providecommand\bibinfo[2]{#2}
\providecommand\natexlab[1]{#1}
\providecommand\showeprint[2][]{arXiv:#2}

\bibitem[Beeferman et~al\mbox{.}(1999)]%
        {beeferman1999statistical}
\bibfield{author}{\bibinfo{person}{Doug Beeferman}, \bibinfo{person}{Adam
  Berger}, {and} \bibinfo{person}{John Lafferty}.}
  \bibinfo{year}{1999}\natexlab{}.
\newblock \showarticletitle{Statistical models for text segmentation}.
\newblock \bibinfo{journal}{\emph{Machine learning}} \bibinfo{volume}{34},
  \bibinfo{number}{1} (\bibinfo{year}{1999}), \bibinfo{pages}{177--210}.
\newblock


\bibitem[Cohan et~al\mbox{.}(2018)]%
        {cohan-etal-2018-discourse}
\bibfield{author}{\bibinfo{person}{Arman Cohan}, \bibinfo{person}{Franck
  Dernoncourt}, \bibinfo{person}{Doo~Soon Kim}, \bibinfo{person}{Trung Bui},
  \bibinfo{person}{Seokhwan Kim}, \bibinfo{person}{Walter Chang}, {and}
  \bibinfo{person}{Nazli Goharian}.} \bibinfo{year}{2018}\natexlab{}.
\newblock \showarticletitle{A Discourse-Aware Attention Model for Abstractive
  Summarization of Long Documents}. In \bibinfo{booktitle}{\emph{Proceedings of
  the 2018 Conference of the North {A}merican Chapter of the Association for
  Computational Linguistics: Human Language Technologies, Volume 2 (Short
  Papers)}}. \bibinfo{publisher}{Association for Computational Linguistics},
  \bibinfo{address}{New Orleans, Louisiana}, \bibinfo{pages}{615--621}.
\newblock
\urldef\tempurl%
\url{https://doi.org/10.18653/v1/N18-2097}
\showDOI{\tempurl}


\bibitem[Devlin et~al\mbox{.}(2019)]%
        {devlin-etal-2019-bert}
\bibfield{author}{\bibinfo{person}{Jacob Devlin}, \bibinfo{person}{Ming-Wei
  Chang}, \bibinfo{person}{Kenton Lee}, {and} \bibinfo{person}{Kristina
  Toutanova}.} \bibinfo{year}{2019}\natexlab{}.
\newblock \showarticletitle{{BERT}: Pre-training of Deep Bidirectional
  Transformers for Language Understanding}. In
  \bibinfo{booktitle}{\emph{Proceedings of the 2019 Conference of the North
  {A}merican Chapter of the Association for Computational Linguistics: Human
  Language Technologies, Volume 1 (Long and Short Papers)}}.
  \bibinfo{publisher}{Association for Computational Linguistics},
  \bibinfo{address}{Minneapolis, Minnesota}, \bibinfo{pages}{4171--4186}.
\newblock
\urldef\tempurl%
\url{https://doi.org/10.18653/v1/N19-1423}
\showDOI{\tempurl}


\bibitem[Grenander et~al\mbox{.}(2019)]%
        {grenander-etal-2019-countering}
\bibfield{author}{\bibinfo{person}{Matt Grenander}, \bibinfo{person}{Yue Dong},
  \bibinfo{person}{Jackie Chi~Kit Cheung}, {and} \bibinfo{person}{Annie
  Louis}.} \bibinfo{year}{2019}\natexlab{}.
\newblock \showarticletitle{Countering the Effects of Lead Bias in News
  Summarization via Multi-Stage Training and Auxiliary Losses}. In
  \bibinfo{booktitle}{\emph{Proceedings of the 2019 Conference on Empirical
  Methods in Natural Language Processing and the 9th International Joint
  Conference on Natural Language Processing (EMNLP-IJCNLP)}}.
  \bibinfo{publisher}{Association for Computational Linguistics},
  \bibinfo{address}{Hong Kong, China}, \bibinfo{pages}{6019--6024}.
\newblock
\urldef\tempurl%
\url{https://doi.org/10.18653/v1/D19-1620}
\showDOI{\tempurl}


\bibitem[Hearst(1997)]%
        {hearst-1997-text}
\bibfield{author}{\bibinfo{person}{Marti~A. Hearst}.}
  \bibinfo{year}{1997}\natexlab{}.
\newblock \showarticletitle{Text Tiling: Segmenting Text into Multi-paragraph
  Subtopic Passages}.
\newblock \bibinfo{journal}{\emph{Computational Linguistics}}
  \bibinfo{volume}{23}, \bibinfo{number}{1} (\bibinfo{year}{1997}),
  \bibinfo{pages}{33--64}.
\newblock
\urldef\tempurl%
\url{https://aclanthology.org/J97-1003}
\showURL{%
\tempurl}


\bibitem[Li et~al\mbox{.}(2018)]%
        {ijcai2018-579}
\bibfield{author}{\bibinfo{person}{Jing Li}, \bibinfo{person}{Aixin Sun}, {and}
  \bibinfo{person}{Shafiq Joty}.} \bibinfo{year}{2018}\natexlab{}.
\newblock \showarticletitle{SegBot: A Generic Neural Text Segmentation Model
  with Pointer Network}. In \bibinfo{booktitle}{\emph{Proceedings of the
  Twenty-Seventh International Joint Conference on Artificial Intelligence,
  {IJCAI-18}}}. \bibinfo{publisher}{International Joint Conferences on
  Artificial Intelligence Organization}, \bibinfo{pages}{4166--4172}.
\newblock
\urldef\tempurl%
\url{https://doi.org/10.24963/ijcai.2018/579}
\showDOI{\tempurl}


\bibitem[Lin(2004)]%
        {lin-2004-rouge}
\bibfield{author}{\bibinfo{person}{Chin-Yew Lin}.}
  \bibinfo{year}{2004}\natexlab{}.
\newblock \showarticletitle{{ROUGE}: A Package for Automatic Evaluation of
  Summaries}. In \bibinfo{booktitle}{\emph{Text Summarization Branches Out}}.
  \bibinfo{publisher}{Association for Computational Linguistics},
  \bibinfo{address}{Barcelona, Spain}, \bibinfo{pages}{74--81}.
\newblock
\urldef\tempurl%
\url{https://aclanthology.org/W04-1013}
\showURL{%
\tempurl}


\bibitem[Liu and Lapata(2019)]%
        {liu-lapata-2019-text}
\bibfield{author}{\bibinfo{person}{Yang Liu} {and} \bibinfo{person}{Mirella
  Lapata}.} \bibinfo{year}{2019}\natexlab{}.
\newblock \showarticletitle{Text Summarization with Pretrained Encoders}. In
  \bibinfo{booktitle}{\emph{Proceedings of the 2019 Conference on Empirical
  Methods in Natural Language Processing and the 9th International Joint
  Conference on Natural Language Processing (EMNLP-IJCNLP)}}.
  \bibinfo{publisher}{Association for Computational Linguistics},
  \bibinfo{address}{Hong Kong, China}, \bibinfo{pages}{3730--3740}.
\newblock
\urldef\tempurl%
\url{https://doi.org/10.18653/v1/D19-1387}
\showDOI{\tempurl}


\bibitem[Liu et~al\mbox{.}(2022)]%
        {liu-etal-2022-end}
\bibfield{author}{\bibinfo{person}{Yang Liu}, \bibinfo{person}{Chenguang Zhu},
  {and} \bibinfo{person}{Michael Zeng}.} \bibinfo{year}{2022}\natexlab{}.
\newblock \showarticletitle{End-to-End Segmentation-based News Summarization}.
  In \bibinfo{booktitle}{\emph{Findings of the Association for Computational
  Linguistics: ACL 2022}}. \bibinfo{publisher}{Association for Computational
  Linguistics}, \bibinfo{address}{Dublin, Ireland}, \bibinfo{pages}{544--554}.
\newblock
\urldef\tempurl%
\url{https://doi.org/10.18653/v1/2022.findings-acl.46}
\showDOI{\tempurl}


\bibitem[Liu et~al\mbox{.}(2019)]%
        {liu2019topic}
\bibfield{author}{\bibinfo{person}{Zhengyuan Liu}, \bibinfo{person}{Angela Ng},
  \bibinfo{person}{Sheldon Lee}, \bibinfo{person}{Ai~Ti Aw}, {and}
  \bibinfo{person}{Nancy~F Chen}.} \bibinfo{year}{2019}\natexlab{}.
\newblock \showarticletitle{Topic-aware pointer-generator networks for
  summarizing spoken conversations}. In \bibinfo{booktitle}{\emph{2019 IEEE
  Automatic Speech Recognition and Understanding Workshop (ASRU)}}. IEEE,
  \bibinfo{pages}{814--821}.
\newblock


\bibitem[Ma et~al\mbox{.}(2021)]%
        {ma2021deltalm}
\bibfield{author}{\bibinfo{person}{Shuming Ma}, \bibinfo{person}{Li Dong},
  \bibinfo{person}{Shaohan Huang}, \bibinfo{person}{Dongdong Zhang},
  \bibinfo{person}{Alexandre Muzio}, \bibinfo{person}{Saksham Singhal},
  \bibinfo{person}{Hany~Hassan Awadalla}, \bibinfo{person}{Xia Song}, {and}
  \bibinfo{person}{Furu Wei}.} \bibinfo{year}{2021}\natexlab{}.
\newblock \showarticletitle{Deltalm: Encoder-decoder pre-training for language
  generation and translation by augmenting pretrained multilingual encoders}.
\newblock \bibinfo{journal}{\emph{arXiv preprint arXiv:2106.13736}}
  (\bibinfo{year}{2021}).
\newblock


\bibitem[Pevzner and Hearst(2002)]%
        {pevzner-hearst-2002-critique}
\bibfield{author}{\bibinfo{person}{Lev Pevzner} {and} \bibinfo{person}{Marti~A.
  Hearst}.} \bibinfo{year}{2002}\natexlab{}.
\newblock \showarticletitle{A Critique and Improvement of an Evaluation Metric
  for Text Segmentation}.
\newblock \bibinfo{journal}{\emph{Computational Linguistics}}
  \bibinfo{volume}{28}, \bibinfo{number}{1} (\bibinfo{year}{2002}),
  \bibinfo{pages}{19--36}.
\newblock
\urldef\tempurl%
\url{https://doi.org/10.1162/089120102317341756}
\showDOI{\tempurl}


\bibitem[Ruan et~al\mbox{.}(2022)]%
        {ruan-etal-2022-histruct}
\bibfield{author}{\bibinfo{person}{Qian Ruan}, \bibinfo{person}{Malte
  Ostendorff}, {and} \bibinfo{person}{Georg Rehm}.}
  \bibinfo{year}{2022}\natexlab{}.
\newblock \showarticletitle{{H}i{S}truct+: Improving Extractive Text
  Summarization with Hierarchical Structure Information}. In
  \bibinfo{booktitle}{\emph{Findings of the Association for Computational
  Linguistics: ACL 2022}}. \bibinfo{publisher}{Association for Computational
  Linguistics}, \bibinfo{address}{Dublin, Ireland},
  \bibinfo{pages}{1292--1308}.
\newblock
\urldef\tempurl%
\url{https://doi.org/10.18653/v1/2022.findings-acl.102}
\showDOI{\tempurl}


\bibitem[Solbiati et~al\mbox{.}(2021)]%
        {solbiati2021unsupervised}
\bibfield{author}{\bibinfo{person}{Alessandro Solbiati}, \bibinfo{person}{Kevin
  Heffernan}, \bibinfo{person}{Georgios Damaskinos}, \bibinfo{person}{Shivani
  Poddar}, \bibinfo{person}{Shubham Modi}, {and} \bibinfo{person}{Jacques
  Cali}.} \bibinfo{year}{2021}\natexlab{}.
\newblock \showarticletitle{Unsupervised Topic Segmentation of Meetings with
  BERT Embeddings}.
\newblock \bibinfo{journal}{\emph{arXiv e-prints}} (\bibinfo{year}{2021}),
  \bibinfo{pages}{arXiv--2106}.
\newblock


\bibitem[Vaswani et~al\mbox{.}(2017)]%
        {10.5555/3295222.3295349}
\bibfield{author}{\bibinfo{person}{Ashish Vaswani}, \bibinfo{person}{Noam
  Shazeer}, \bibinfo{person}{Niki Parmar}, \bibinfo{person}{Jakob Uszkoreit},
  \bibinfo{person}{Llion Jones}, \bibinfo{person}{Aidan~N. Gomez},
  \bibinfo{person}{\L{}ukasz Kaiser}, {and} \bibinfo{person}{Illia
  Polosukhin}.} \bibinfo{year}{2017}\natexlab{}.
\newblock \showarticletitle{Attention is All You Need}. In
  \bibinfo{booktitle}{\emph{Proceedings of the 31st International Conference on
  Neural Information Processing Systems}} (Long Beach, California, USA)
  \emph{(\bibinfo{series}{NIPS'17})}. \bibinfo{publisher}{Curran Associates
  Inc.}, \bibinfo{address}{Red Hook, NY, USA}, \bibinfo{pages}{6000–6010}.
\newblock
\showISBNx{9781510860964}


\bibitem[Xu et~al\mbox{.}(2021)]%
        {xu2021topicaware}
\bibfield{author}{\bibinfo{person}{Yi Xu}, \bibinfo{person}{Hai Zhao}, {and}
  \bibinfo{person}{Zhuosheng Zhang}.} \bibinfo{year}{2021}\natexlab{}.
\newblock \showarticletitle{Topicaware multi-turn dialogue modeling}. In
  \bibinfo{booktitle}{\emph{The Thirty-Fifth AAAI Conference on Artificial
  Intelligence (AAAI-21)}}.
\newblock


\bibitem[Zhang et~al\mbox{.}(2021)]%
        {zhang2021sequence}
\bibfield{author}{\bibinfo{person}{Qinglin Zhang}, \bibinfo{person}{Qian Chen},
  \bibinfo{person}{Yali Li}, \bibinfo{person}{Jiaqing Liu}, {and}
  \bibinfo{person}{Wen Wang}.} \bibinfo{year}{2021}\natexlab{}.
\newblock \showarticletitle{Sequence Model with Self-Adaptive Sliding Window
  for Efficient Spoken Document Segmentation}.
\newblock \bibinfo{journal}{\emph{arXiv preprint arXiv:2107.09278}}
  (\bibinfo{year}{2021}).
\newblock


\end{thebibliography}


\end{document}